\documentclass{article}

\usepackage{arxiv}

\usepackage[utf8]{inputenc} 
\usepackage[T1]{fontenc}    
\usepackage{url}            
\usepackage{booktabs}       
\usepackage{amsfonts}       
\usepackage{nicefrac}       
\usepackage{microtype}      
\usepackage{graphicx}
\usepackage{natbib}
\usepackage{doi}
\usepackage{lmodern}
\usepackage{amsmath,amssymb,amsthm}
\usepackage{caption}
\usepackage{subcaption}
\usepackage{multirow}
\usepackage{array}
\usepackage{rotating}
\usepackage{cleveref}
\usepackage{adjustbox}
\usepackage{makecell}
\usepackage{threeparttable}

\hypersetup{hidelinks}

\graphicspath{{figures/}{../figures/}{paper/figures/}{NEW/paper/figures/}}

\title{AgentFloor: How Far Up the tool use Ladder Can Small Open-Weight Models Go?}

\author{ \href{https://orcid.org/0000-0000-0000-0000}{\includegraphics[scale=0.06]{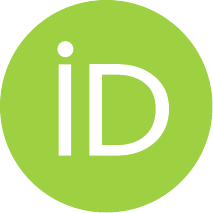}\hspace{1mm}Ranit Karmakar} \\
	Harvard University, 
	Boston, MA 02115 \\
	\texttt{ranit.karmakar@outlook.com} \\
	\And
	\href{https://orcid.org/0009-0001-1814-258X}{\includegraphics[scale=0.06]{orcid.pdf}\hspace{1mm}Jayita Chatterjee} \\
	Boston, MA 02115 \\
	\texttt{jayita\_chatterjee@outlook.com} \\
}

\renewcommand{\shorttitle}{AgentFloor: A Capability-Threshold Benchmark for Agentic Tool Use}

\hypersetup{
pdftitle={A template for the arxiv style},
pdfsubject={q-bio.NC, q-bio.QM},
pdfauthor={David S.~Hippocampus, Elias D.~Striatum},
pdfkeywords={First keyword, Second keyword, More},
}

\begin{document}
\maketitle

\begin{abstract}
Production agentic systems make many model calls per user request, and most of those calls are short, structured, and routine. This raises a practical routing question that existing evaluations do not directly answer: which parts of an agent workflow truly require large frontier intelligence, and which can be handled by smaller models? We introduce AgentFloor, a deterministic 30-task benchmark organized as a six-tier capability ladder, spanning instruction following, tool use, multi-step coordination, and long-horizon planning under persistent constraints. We evaluate 16 open-weight models, from 0.27B to 32B parameters, alongside GPT-5 across 16,542 scored runs. Our results reveal a clear boundary of model necessity. Small and mid-sized open-weight models are already sufficient for much of the short-horizon, structured tool use work that dominates real agent pipelines, and in aggregate, the strongest open-weight model matches GPT-5 on our benchmark while being substantially cheaper and faster to run. The gap appears most clearly on long-horizon planning tasks that require sustained coordination and reliable constraint tracking over many steps, where frontier models still hold an advantage, though neither side reaches strong reliability. We also find that this boundary is not explained by scale alone: some failures respond to targeted interventions, but the effects are model-specific rather than universal. These findings suggest a practical design principle for agentic systems: use smaller open-weight models for the broad base of routine actions, and reserve large frontier models for the narrower class of tasks that truly demand deeper planning and control. We release the benchmark, harness, sweep configurations, and full run corpus.
\end{abstract}

\section{Introduction}
\label{sec:intro}

Agentic LLM systems often make many model calls for a single user-visible action, and a substantial fraction of those calls are short, structured, and operationally simple. In production, they frequently take the form of a search, a lookup, a record extraction, or a single submission. The common deployment default is to route all such calls to a frontier flagship model. The practical systems question is whether that default is necessary, and if not, where the routing boundary should lie. Concretely, how far up the cognitive capabilities can small models go before a frontier model becomes the justified choice?

Existing evaluations do not answer this question directly. Single-turn function-calling benchmarks such as BFCL, API-Bank, and Gorilla measure isolated tool use, but they abstract away the sequential dependencies that dominate cost in deployed systems. Multi-step agentic suites such as ToolBench, $\tau$-bench, MINT, AgentBench, AgentBoard, GAIA, WebArena, OSWorld, and SWE-Bench are closer to real workflows, but they also entangle core tool use ability with confounds such as API drift, web or GUI grounding, and possible contamination. Capability-scaling studies typically relate aggregate performance to model size without decomposing the distinct cognitive demands within multi-step tool use. Prompting studies vary scaffolding strategies, but generally do not connect intervention effects to a controlled capability ladder. Failure-taxonomy work clarifies how agents break down, and routing work clarifies why model selection matters, yet neither yields a controlled map of where open-weight models match frontier systems and where they do not.

We introduce AgentFloor, a deterministic 30-task benchmark organized as a six-tier capability ladder over eight abstract tools and an in-memory fixture database. The tiers isolate progressively harder demands: instruction following without tools, single-tool invocation, sequential two-tool chaining, conditional branching on an intermediate result, multi-source synthesis with conflict recovery, and long-horizon planning under persistent constraints. All tasks run in a fixed abstract-tool environment with no filesystem, no live APIs, no time-varying state, and no plausible route to pretraining-corpus contamination. This design narrows the evaluation target. Rather than asking whether agents succeed in open environments, we ask where models of different scales reliably succeed as tool users once environmental noise is removed and cognitive demands are explicitly tiered.

\begin{figure}[tb]
  \centering
  \includegraphics[width=\linewidth]{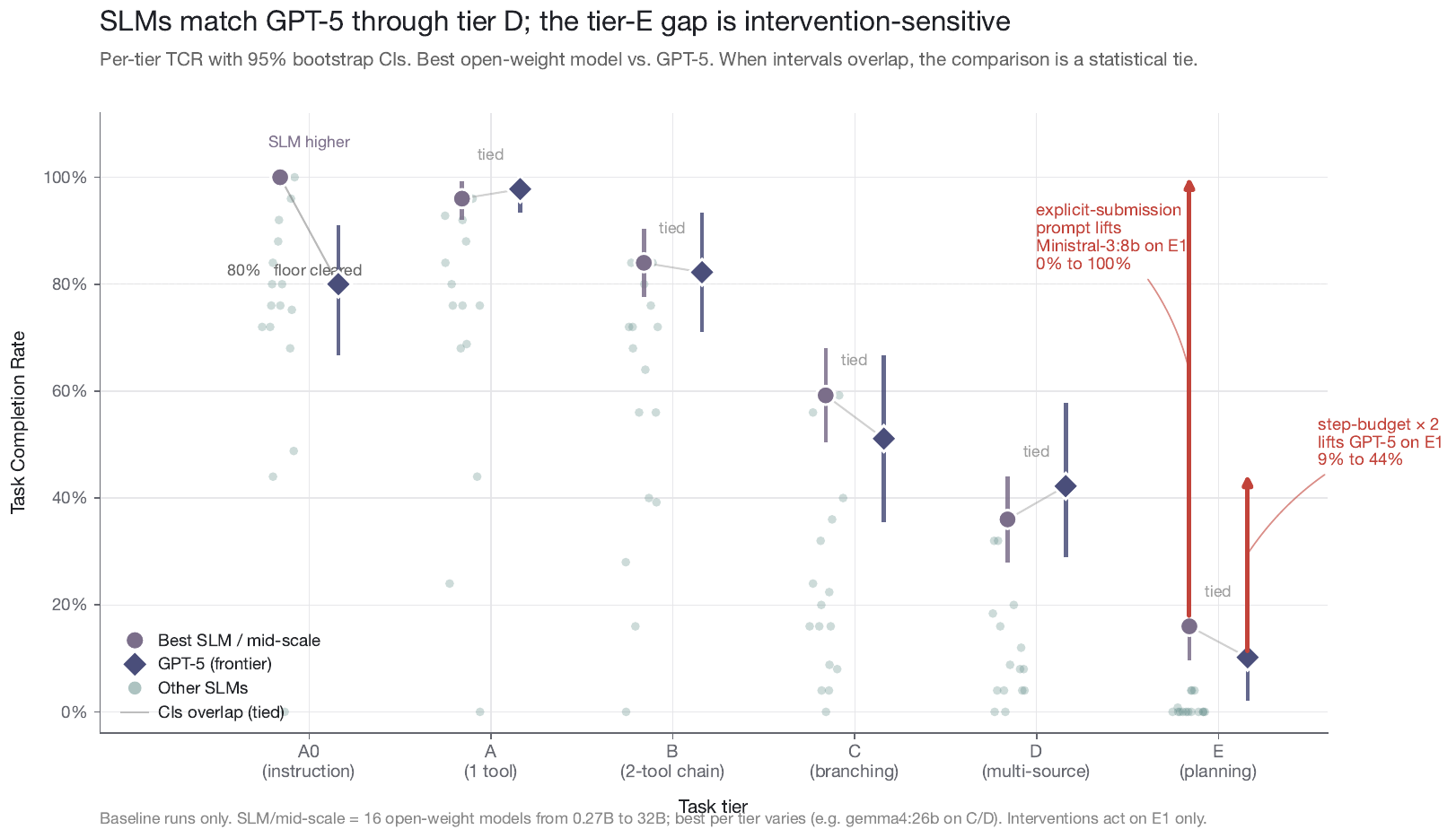}
  \caption{The AgentFloor capability ladder. Six tiers (A0--E), each introducing one new cognitive demand. Per-tier TCR shown for \textit{gemma4:26b} (the Frame~A open-weight anchor; highest overall TCR among open-weight models in our corpus) and GPT-5. Per-tier-best open-weight cells differ by tier; see \Cref{tab:tcr} for the full breakdown.}
  \label{fig:ladder}
\end{figure}

We evaluate 16 open-weight models spanning 0.27B to 32B parameters alongside GPT-5 family using native tool calling only, paired bootstrap analyses, and a strict failure-mode cascade. The resulting picture is stratified rather than uniform. At the aggregate level, the best open-weight model in our corpus, gemma4:26b, is equivalent to GPT-5 within a pre-registered margin while operating at substantially lower cost and lower latency per passed task. At the tier level, open-weight models outperform GPT-5 on no-tool instruction following and are formally equivalent on single-tool use. On sequential chaining, branching, and multi-source synthesis, point estimates remain close to GPT-5, although the per-tier paired sample is not large enough to certify equivalence at the pre-registered margin. The clearest frontier advantage appears on long-horizon planning under persistent constraints. Even there, however, neither side reaches reliability that would support confident deployment.

These results matter for routing. They suggest that a large share of production agent calls can be handled by small or mid-scale open-weight models without measurable loss on lower-complexity tiers, while the residual gap is concentrated in a narrower long-horizon regime. They also show that this gap is not explained by a single universal intervention. In our ablations, interventions that help one model are typically null on others, while a seemingly natural structured decomposition prompt degrades performance across all tested models. AgentFloor therefore contributes not only a benchmark, but also a capability-and-cost map and a failure-mode account of where routing decisions should change.

Our contributions are threefold. First, we introduce AgentFloor, a deterministic six-tier benchmark for controlled evaluation of tool use capability. Second, we provide a pre-registered capability-and-cost comparison across 16 open-weight models and GPT-5. Third, we analyze the residual long-horizon gap through targeted ablations and tier-specific failure signatures. Taken together, these results provide a concrete answer to a practical deployment question: which parts of the tool use stack can already be delegated to open-weight models, and which parts still justify frontier routing?

\section{Related Work}
\label{sec:related}

\paragraph{Tool use and agentic benchmarks.}
Single-turn function-calling benchmarks~\citep{patil2024bfcl, li2023apibank, patil2023gorilla} and multi-step agentic suites~\citep{guo2024stabletoolbench,yao2024taubench,wang2023mint,liu2023agentbench,ma2024agentboard,mialon2023gaia,zhou2023webarena,xie2024osworld,jimenez2023swe} cover much of the space, but most report aggregate scores without bootstrap CIs and few pair results with a structured failure-mode taxonomy. The closest neighbours are $\tau$-bench~\citep{yao2024taubench} (\textit{pass\textasciicircum k} reliability), API-Bank~\citep{li2023apibank} (three ability levels by tool count), AgentBoard~\citep{ma2024agentboard} (progress rate plus a sub-skill panel), and ComplexFuncBench~\citep{zhong2025complexfuncbench} (which tracks \textit{stop\_early} as a failure category). API drift on real-API benchmarks (ToolBench's 44.4\% API success at rerun) and contamination on SWE-Bench Verified (59.4\% of audited problems with flawed test cases per OpenAI's 2025 reanalysis) motivate our deterministic abstract-tool design; the closest precedents are StableToolBench~\citep{guo2024stabletoolbench}, ToolEmu~\citep{ruan2023toolemu}, and REAL~\citep{garg2025real}.

\paragraph{SLM capability and scaling.}
SLM technical reports rarely include agentic evaluation~\citep{abdin2024phi3,yang2025qwen3,gemma2024report,abdelaziz2024granite,mahmood2025smallmodelsbigtasks,grattafiori2024llama}. Capability surveys~\citep{lu2024slmsurvey} and concurrent SLM tool use studies~\citep{mahmood2025smallmodelsbigtasks} measure adjacent quantities --- JSON validity, qualitative failure archetypes --- but do not produce a capability map across a controlled cognitive-demand ladder. Position papers~\citep{belcak2025slmsfutureagentic} argue that small models are sufficient for agentic deployment; we quantify the claim.

\paragraph{Failure-mode taxonomies.}
ToolScan/SpecTool~\citep{kokane2024toolscan}, CriticTool~\citep{yuan2025critictool}, and MAST~\citep{cemri2025mast} catalogue agentic failures. None pairs a tier-graded failure-mode comparison with a quantitative capability map.

\paragraph{Cost-quality routing.}
FrugalGPT~\citep{chen2023frugalgpt}, RouteLLM~\citep{ong2024routellm}, Hybrid LLM~\citep{ding2024hybridllm}, and AutoMix~\citep{aggarwal2023automix} route between models by predicted difficulty. The capability-and-cost map below provides a static prior these systems require: tiers A0/A/B can be routed to a sub-5\,B open-weight model at no measurable accuracy cost in our corpus; the residual gap on E is intervention-sensitive (\Cref{sec:interventions}) rather than scale-bound.

\section{Methods}
\label{sec:methods}

\paragraph{Capability ladder.}
AgentFloor consists of 30 tasks across six tiers (A0/A/B/C/D/E), 5 tasks per tier. Each tier introduces one new cognitive demand: instruction following without tools (A0), single tool call (A), sequential 2-tool chain where the first tool's output feeds the second (B), branch on intermediate result (C), multi-source synthesis with conflict recovery (D), long-horizon planning under persistent constraints (E). Step budgets are 1/2/4/6/8/10. Each task ships with a canonical prompt (v0) and four paraphrase variants (v1--v4); four tasks (A1, B1, C1, E1) additionally have five instance variants (i1--i5) for the instance-variation ablation.

\paragraph{Tool surface.}
All tasks are solved through the same eight deterministic tools (\textit{search\_records}, \textit{lookup\_record}, \textit{get\_attribute}, \textit{list\_options}, \textit{check\_constraint}, \textit{compare\_records}, \textit{compute\_value}, \textit{submit\_decision}) over an in-memory fixture database. Per-task fixtures parameterise tool behaviour without changing the schema. There is no filesystem, no external service, no real timestamp, and no plausible route to pretraining-corpus contamination.

\paragraph{Inference protocol.}
A single Python runner controls every run with no model-specific prompts and no provider-specific repair logic. Tool calls are read only from each provider's \emph{native} tool-calling field; text-extracted JSON is treated as zero tool calls. Models that probe as \textit{text\_leaked}, \textit{no\_tools}, or \textit{error} are excluded by design; the benchmark measures native tool-calling control, not prompt-based emulation. Temperature is 0; the system prompt is byte-identical across all models and conditions, except for the structured-prompt and explicit-submission ablations.

\paragraph{Scoring.}
TCR is binary pass/fail per run. A run passes iff four checker families return pass: final-answer fields match the gold state; the submission validator accepts the payload (where applicable); the trajectory satisfies the required tool sequence and per-task predicates; and no forbidden behaviour fires (hallucinated tool, terminate-without-answer, repeated-identical-call, partial-constraint-check). Three semantic predicates (A02/A03/A05 hallucinated-facts, E5 inconsistent-recovery) are routed through an LLM judge (\textit{gpt-5-nano}, deterministic, SHA-256-cached, $\approx 88\%$ cache hit rate). Bootstrap CIs at 95\% use $n_{\textrm{boot}}=10{,}000$, seed 42, with (variant, run\_idx) tuples as resampling units.

\paragraph{Frontier comparisons.}
We report two-one-sided-test (TOST) equivalence and one-sided non-inferiority results at a pre-registered $\pm 10$\,pp margin. The pre-registered Frame~A anchors \textit{gemma4:26b} (the highest overall TCR among open-weight models in our corpus) against GPT-5; the exploratory Frame~B compares all 16 open-weight models against GPT-5 across all six tiers (96 paired tests) with Holm--Bonferroni correction at family-wise $\alpha = 0.05$. Pairing is on (task, variant, run\_idx) tuples; the inner-join restricts to GPT-5's 3 variants $\times$ 3 runs $\times$ 5 tasks $= 45$ paired observations per tier on A0/A/B/C/D and 49 on E (additional GPT-5 baseline runs in the explicit-submission ablation), giving 270 paired observations across the canonical 30 tasks. Equivalence claims use the 90\% CI of $\Delta$ (per the standard TOST construction); descriptive intervals on $\Delta$ in tables use 95\%, except where the bootstrap support is discrete enough that 90\% and 95\% intervals coincide.

\paragraph{Failure taxonomy.}
Failing runs are classified by a strict priority cascade into F1 (hallucinated tool), F2 (malformed call), F4 (step-budget exhausted), F5 (early resignation: $\geq 2$ successful tool calls then quit), F5b (plan-without-execute: $\leq 1$ tool call), F6 (wrong tool), F7 (partial completion), F3 (residual). The F5/F5b split distinguishes engaged-then-resigned from never-engaged failures; it carries the \Cref{sec:failures} mechanism analysis.

\paragraph{Corpus and pricing.}
16 open-weight models from 0.27\,B to 32\,B served via Ollama, plus GPT-5 via the OpenAI Responses API. The main sweep contributes 12{,}000 SLM runs and 274 GPT-5 anchor runs; ablations contribute the remainder (instance variation, Qwen3 reasoning toggle, structured-prompt, GPT-5 step-budget $\times 2$, E-tier explicit-submission). Total: 16{,}542 scored runs; full inventory. Pricing follows locked 2026-04 list rates: Mac self-hosted at \$0.50/hr amortised; cloud GPU tiered at \$0.50/\$1.50/\$2.50 per hour ($\leq 4$\,B / 7--14\,B / $\geq 20$\,B); GPT-5 at \$1.25/\$10.00 per million in/out tokens.

\section{Results}
\label{sec:results}

\subsection{Aggregate parity and the cost picture}
\label{sec:parity}

Across all 30 tasks paired between \textit{gemma4:26b} and GPT-5 ($n = 270$ paired observations), the two are equivalent at the pre-registered $\pm 10$\,pp margin (paired $\Delta = +0.4$\,pp, 90\% CI $[-4.0, +5.1]$; the same data certify equivalence at any margin $\geq 5.1$\,pp). This is the most statistically powered comparison in the corpus and the parity claim that anchors the cost analysis below. The per-tier breakdown in \Cref{sec:tierby} sharpens it.

The cost consequence is sizeable. At matched aggregate accuracy ($\sim 60\%$ overall TCR for both models), \textit{gemma4:26b} on a Mac Studio amortised at \$0.50/hr reaches \$0.0022 per passed task, against GPT-5 at \$0.0327 at posted prices --- about 15$\times$ cheaper. On cloud GPU at \$2.50/hr (H100 spot), \textit{gemma4:26b} is about 3$\times$ cheaper than GPT-5 per passed task. Smaller open-weight models reach lower aggregate accuracy at proportionally lower cost: \textit{granite4:3b} at 40\% aggregate TCR sits at \$0.00046 per passed task on Mac, 71$\times$ cheaper than GPT-5. Wall-clock latency follows the same direction: at matched aggregate accuracy, \textit{gemma4:26b} runs 2.5$\times$ faster per passed task. The Pareto frontier of cost-per-passed-task is occupied by open-weight models at every reliability point in our corpus.

\subsection{The tier-by-tier breakdown}
\label{sec:tierby}

\begin{figure}[tb]
  \centering
  \includegraphics[width=\linewidth]{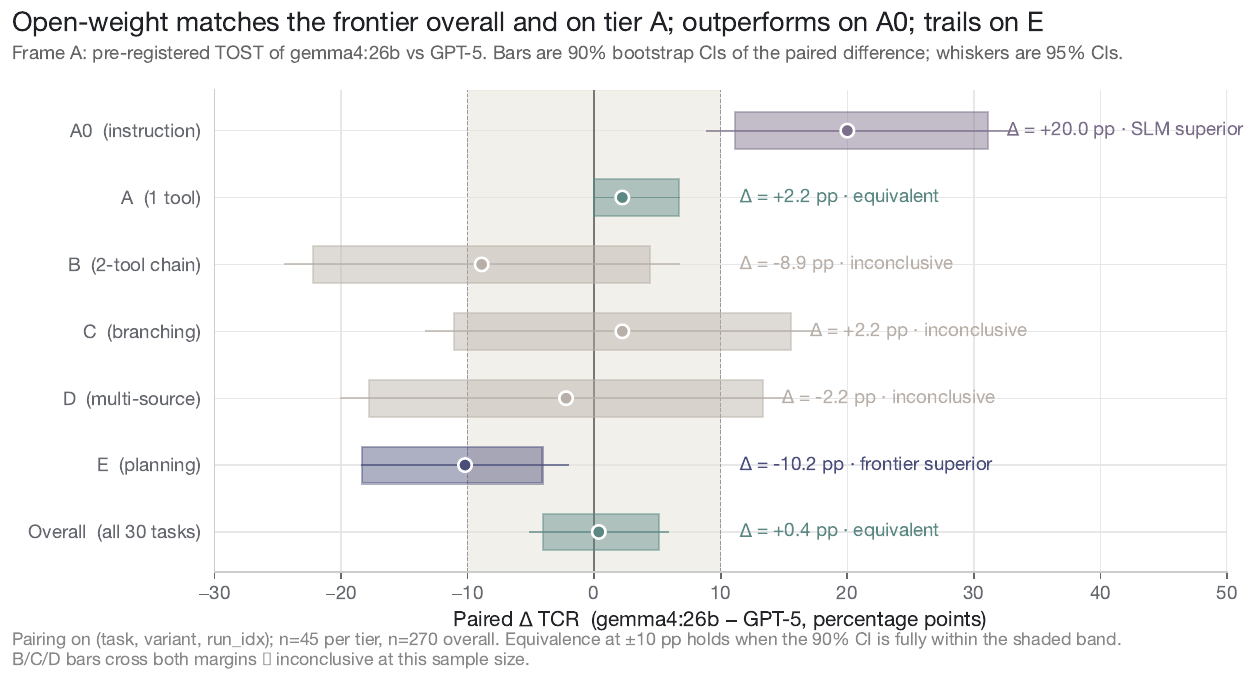}
  \caption{Frame~A paired difference $\Delta = $ \textit{gemma4:26b} $-$ GPT-5 with 90\% bootstrap CIs per tier (and overall, paired across all 30 tasks). The shaded band marks the pre-registered $\pm 10$\,pp equivalence margin. The 90\% CI for A0 sits strictly above the upper margin (open-weight strict superiority); A and the overall comparison have CIs wholly within the margin (TOST equivalence); E has a CI strictly below the lower margin (frontier strict superiority). On B, C, and D the per-tier 90\% CI extends past the margin on at least one side, so neither equivalence nor inferiority at $\pm 10$\,pp is formally established at this sample size.}
  \label{fig:equivalence}
\end{figure}

\Cref{tab:tost} reports the Frame~A paired TOST per tier; \Cref{fig:equivalence} visualises the same data with the equivalence margin overlaid; \Cref{tab:tcr} reports the full per-model per-tier TCR with 95\% bootstrap CIs. The verdict varies by tier.

\begin{table}[tb]
  \centering
  \footnotesize
  \caption{Frame~A: pre-registered paired TOST of \textit{gemma4:26b} vs GPT-5. $\Delta$ = mean(\textit{gemma4:26b}) $-$ mean(GPT-5). Equivalence at $\pm 10$\,pp holds if 90\% CI of $\Delta$ is within $[-10, +10]$\,pp. Non-inferiority holds if 95\% CI lower bound of $\Delta$ is $> -10$\,pp. Frame~B Holm passes counts SLMs (out of 16) that pass non-inferiority on this tier under family-wise $\alpha=0.05$.}
  \label{tab:tost}
  \begin{tabular}{lrrrrrccc}
    \toprule
    Tier & $n_\text{pairs}$ & \textit{gemma4:26b} & GPT-5 & $\Delta$ (pp) & 95\% CI of $\Delta$ & Equiv. $\pm 10$ & Non-inf. $\pm 10$ & Frame~B Holm \\
    \midrule
    A0       & 45  & 100.0\% & 80.0\% & \textbf{+20.0} & $[+8.9, +33.3]$  & no & \textbf{yes}         & 5 SLMs \\
    A        & 45  & 100.0\% & 97.8\% & $+2.2$         & $[0.0, +6.7]$    & \textbf{yes}         & \textbf{yes}                    & 3 SLMs \\
    B        & 45  & 73.3\%  & 82.2\% & $-8.9$         & $[-24.4, +6.7]$  & no                   & no                              & 0      \\
    C        & 45  & 53.3\%  & 51.1\% & $+2.2$         & $[-13.3, +17.8]$ & no                   & no                              & 0      \\
    D        & 45  & 40.0\%  & 42.2\% & $-2.2$         & $[-20.0, +15.6]$ & no                   & no                              & 0      \\
    E        & 49  & 0.0\%   & 10.2\% & \textbf{$-10.2$} & $[-18.4, -2.0]$ & no                   & no & 0      \\
    \midrule
    Overall  & 270 & 60.0\%  & 59.6\% & $+0.4$         & $[-5.1, +5.8]$   & \textbf{yes}         & \textbf{yes}                    & ---    \\
    \bottomrule
  \end{tabular}
\end{table}

On no-tool instruction following (A0), \textit{gemma4:26b} strictly outperforms GPT-5: 100\% vs 80\%, paired $\Delta = +20.0$\,pp (95\% CI $[+8.9, +33.3]$). Five SLMs Holm-pass non-inferiority on this tier (\textit{gemma4:26b}, \textit{mistral-small3.2:24b}, \textit{nemotron-3-nano:4b}, \textit{qwen3:14b}, \textit{qwen3:8b}). All nine of GPT-5's A0 failures land on a single task --- A02, an incident-summary element-coverage task --- and the dominant failure mode is F5: the model paraphrases the source passage using technical vocabulary that the keyword matcher does not recognise, then declines to revise. The same harness scores \textit{gemma4:26b} at 100\% and \textit{granite4:3b} (3\,B) at 84\% [78, 90] on the same task, so this is not a harness artifact. It is a frontier model under-using a task its smaller peers complete reliably.

On single-tool use (A), \textit{gemma4:26b} is formally equivalent to GPT-5 at the $\pm 10$\,pp margin: $\Delta = +2.2$\,pp, 95\% CI $[0, +6.7]$ (the 95\% interval is wholly within $\pm 10$\,pp, so the narrower 90\% interval used for TOST is also within). Three SLMs Holm-pass non-inferiority on this tier.

On B (sequential chains), C (branching), and D (multi-source synthesis with recovery), point estimates differ from GPT-5 by less than 9\,pp on all three tiers, but per-tier paired sample size ($n = 45$) yields 95\% CIs on the difference of width 24--35\,pp --- wider than the pre-registered $\pm 10$\,pp margin. The data on these tiers are not informative enough to certify equivalence at this margin, and we do not claim per-tier equivalence on B, C, or D. We report the point estimates and intervals descriptively; a larger paired frontier sample would be expected to tighten them.

On long-horizon planning under persistent constraints (E), GPT-5 is strictly superior to \textit{gemma4:26b}: $\Delta = -10.2$\,pp, 95\% CI $[-18.4, -2.0]$. Absolute pass rates are 10\% for GPT-5 and 0\% for \textit{gemma4:26b} --- neither side reaches practitioner-relevant reliability on E. The highest open-weight cell on this tier is \textit{ministral-3:8b} at 16\% (CI overlapping GPT-5's), but it does not Holm-pass non-inferiority at any margin in our sweep.

\begin{table}[tb]
  \centering
  \footnotesize
  \caption{Per-tier TCR with 95\% bootstrap CIs. Cells are point estimate\% [CI low, CI high]. \textbf{Bold} = best per tier. $n=125$ per (model, tier) cell for the SLM corpus; $n \geq 45$ for GPT-5. Bootstrap: 10{,}000 resamples, seed 42.}
  \label{tab:tcr}
  \begin{tabular}{llrrrrrr}
    \toprule
    Model & Params & A0 & A & B & C & D & E \\
    \midrule
    \textit{functiongemma:270m}    & 0.27\,B & 0\,[0,0]                & 0\,[0,0]                & 0\,[0,0]                & 0\,[0,0]                & 0\,[0,0]                & 0\,[0,0] \\
    \textit{qwen3:0.6b}            & 0.6\,B  & 80\,[73,86]             & 44\,[35,53]             & 39\,[30,48]             & 4\,[1,8]                & 8\,[4,13]               & 0\,[0,0] \\
    \textit{qwen3:1.7b}            & 1.7\,B  & 76\,[68,83]             & 69\,[61,77]             & 40\,[31,49]             & 9\,[4,14]               & 8\,[3,13]               & 0\,[0,0] \\
    \textit{qwen3.5:2b}            & 2\,B    & 44\,[35,53]             & 68\,[60,76]             & 80\,[73,86]             & 40\,[31,49]             & 32\,[24,40]             & 0\,[0,0] \\
    \textit{granite4:3b}           & 3\,B    & 84\,[78,90]             & 80\,[73,87]             & 56\,[47,65]             & 16\,[10,22]             & 4\,[1,8]                & 0\,[0,0] \\
    \textit{ministral-3:3b}        & 3\,B    & 72\,[64,80]             & 88\,[82,94]             & 56\,[47,65]             & 16\,[10,22]             & 20\,[14,27]             & 4\,[1,8] \\
    \textit{gemma4:e4b}            & 4\,B    & 68\,[60,76]             & 96\,[92,99]             & 76\,[69,83]             & 32\,[24,40]             & 4\,[1,8]                & 0\,[0,0] \\
    \textit{nemotron-3-nano:4b}    & 4\,B    & 92\,[87,96]             & 76\,[68,83]             & 72\,[64,80]             & 20\,[14,27]             & 36\,[28,44]             & 0\,[0,0] \\
    \textit{ministral-3:8b}        & 8\,B    & 75\,[67,82]             & 84\,[78,90]             & \textbf{84\,[78,90]}    & 56\,[47,65]             & 16\,[10,22]             & \textbf{16\,[10,22]} \\
    \textit{qwen3:8b}              & 8\,B    & 80\,[73,87]             & 76\,[68,83]             & 64\,[55,72]             & 24\,[17,32]             & 0\,[0,0]                & 0\,[0,0] \\
    \textit{ministral-3:14b}       & 14\,B   & 49\,[40,58]             & 24\,[17,32]             & 28\,[20,36]             & 4\,[1,8]                & 4\,[1,8]                & 4\,[1,8] \\
    \textit{qwen3:14b}             & 14\,B   & 88\,[82,94]             & 92\,[87,97]             & 84\,[78,90]             & 16\,[10,22]             & 4\,[1,8]                & 4\,[1,8] \\
    \textit{gpt-oss:20b}           & 20\,B   & 72\,[64,80]             & 76\,[68,83]             & 68\,[59,76]             & 22\,[15,30]             & 18\,[12,26]             & 1\,[0,2] \\
    \textit{mistral-small3.2:24b}  & 24\,B   & 96\,[92,99]             & 93\,[88,97]             & 16\,[10,22]             & 8\,[4,14]               & 12\,[6,18]              & 0\,[0,0] \\
    \textit{gemma4:26b}            & 26\,B   & \textbf{100\,[100,100]} & 96\,[92,99]             & 72\,[64,80]             & \textbf{59\,[50,68]}    & 32\,[24,40]             & 0\,[0,0] \\
    \textit{qwen3:32b}             & 32\,B   & 76\,[68,83]             & 96\,[92,99]             & 72\,[64,80]             & 36\,[28,45]             & 9\,[4,14]               & 0\,[0,0] \\
    GPT-5                          & ---     & 80\,[67,91]             & \textbf{98\,[93,100]}   & 82\,[71,93]             & 51\,[36,67]             & \textbf{42\,[29,58]}    & 10\,[2,18] \\
    \bottomrule
  \end{tabular}
\end{table}

\begin{figure}[ht!]
  \centering
  \includegraphics[width=\linewidth]{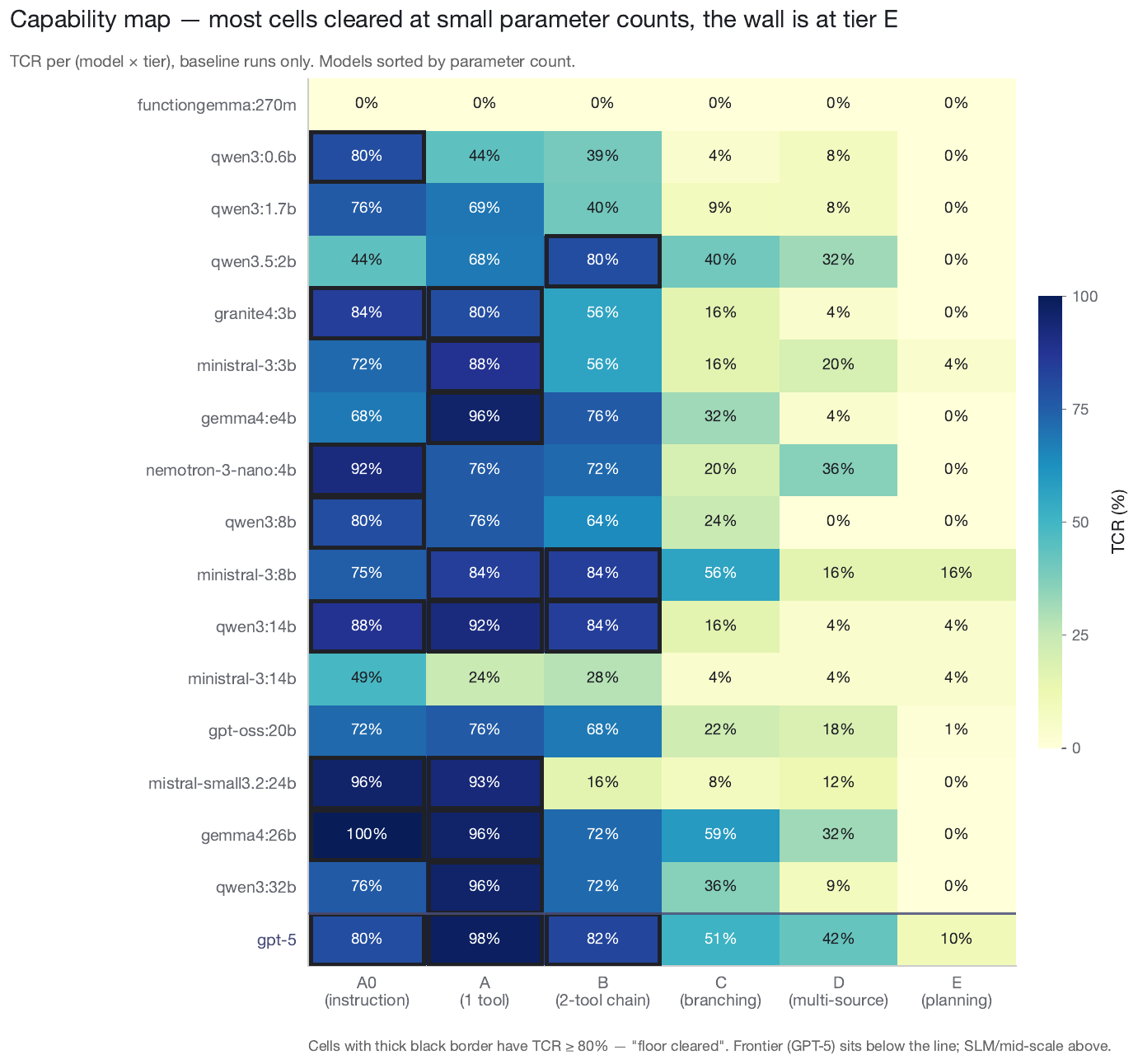}
  \caption{Capability heatmap: 17 models $\times$ 6 tiers, cell value is TCR\%. Most open-weight models reach high TCR on A0 and A across the 3--26\,B range. The B$\rightarrow$C transition is the steepest column-step in the corpus. The GPT-5 row tracks the \textit{gemma4:26b} row closely except on A0 (\textit{gemma4} higher) and E (GPT-5 higher).}
  \label{fig:heatmap}
\end{figure}

The capability heatmap (\Cref{fig:heatmap}) compresses \Cref{tab:tcr} into a single visual. Three features stand out. Most open-weight models reach high TCR on A0 and A across the 3--26\,B range. The B$\rightarrow$C transition is the steepest single column-step in the heatmap. The GPT-5 row tracks the \textit{gemma4:26b} row closely at every tier except A0 (\textit{gemma4} higher) and E (GPT-5 higher). Three rows are conspicuously dim and worth flagging: \textit{functiongemma:270m} as the corpus floor, \textit{ministral-3:14b} which is dominated cell by cell by its 8\,B sibling, and \textit{mistral-small3.2:24b} on B at 16\% in an otherwise strong row.

\subsection{The smallest open-weight model that's reliable enough}
\label{sec:smallest}

\begin{figure}[tb]
  \centering
  \includegraphics[width=\linewidth]{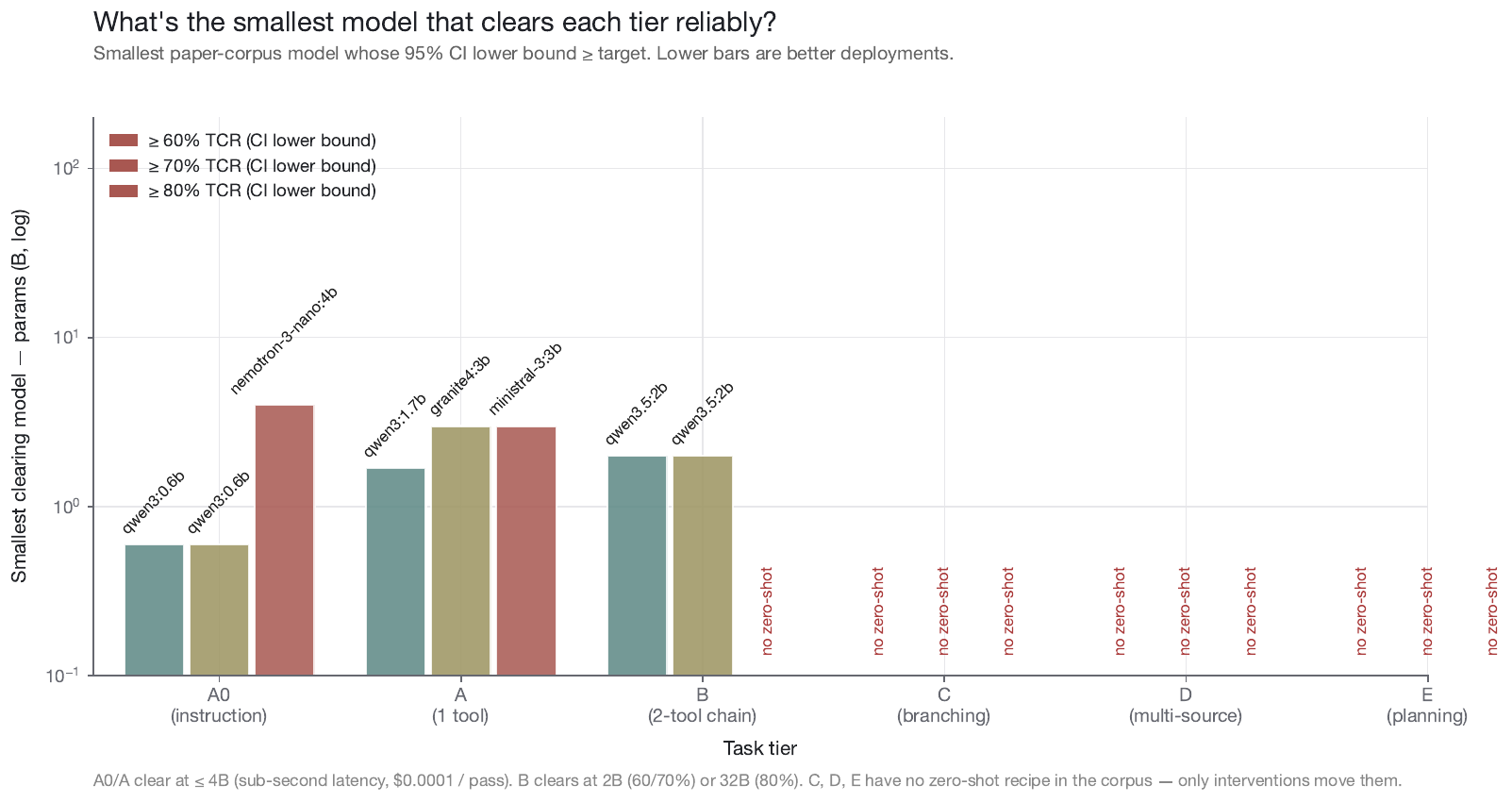}
  \caption{Smallest open-weight model per (tier, threshold) cell whose 95\% bootstrap CI lower bound clears the reliability target. A0, A, and B are cleared by sub-5\,B models at the 70\%--80\% targets in this corpus. No model in this corpus clears any threshold in $[60, 90]$ on C, D, or E without intervention.}
  \label{fig:smallest}
\end{figure}

For each tier and each reliability target $\tau \in \{60, 70, 80, 90\}$, \Cref{fig:smallest} plots the smallest open-weight model in our corpus whose 95\% CI lower bound clears $\tau$. The 95\%-CI-lower-bound criterion is strict: a model is credited as clearing $\tau$ only if the bootstrap rules out the possibility that its true TCR is below $\tau$.

At the standard 80\% bar, A0 clears at 4\,B (\textit{nemotron-3-nano:4b}, 92\% [87, 96]); A clears at 3\,B (\textit{ministral-3:3b}, 88\% [82, 94]); B does not clear at any size in our corpus --- the tightest tier-B cells are \textit{ministral-3:8b} and \textit{qwen3:14b}, both at 84\% [78, 90], two points below the bar. Relaxing to 70\% admits a 2\,B model on B (\textit{qwen3.5:2b}, 80\% [73, 86]); relaxing to 60\% admits a 0.6\,B model on A0 (\textit{qwen3:0.6b}, 80\% [73, 86]). C, D, and E never clear at any threshold in $[60, 90]$. The cliff between B and C is therefore a qualitative break in the corpus that survives any reasonable threshold choice within that range, not an artifact of the 80\% convention.

\begin{table}[tb]
  \centering
  \footnotesize
  \caption{Deployment recipes: smallest model whose 95\% CI lower bound clears the reliability target on each tier. ``---'' = no zero-shot model in the corpus clears the target. Cost and latency are best-cell values on Mac self-hosted.}
  \label{tab:recipes}
  \begin{tabular}{lllllr}
    \toprule
    Tier & 60\% target & 70\% target & 80\% target & Cost/pass (best) & Latency/pass (best) \\
    \midrule
    A0 & \textit{qwen3:0.6b}  & \textit{qwen3:0.6b} & \textit{nemotron-3-nano:4b} & \$0.0002  & 1.4\,s \\
    A  & \textit{qwen3:1.7b}  & \textit{granite4:3b}  & \textit{ministral-3:3b}     & \$0.00007 & 0.5\,s \\
    B  & \textit{qwen3.5:2b}    & \textit{qwen3.5:2b}   & ---                                & \$0.0010  & 7.3\,s \\
    C  & ---                           & ---                           & ---                                & ---       & ---     \\
    D  & ---                           & ---                           & ---                                & ---       & ---     \\
    E  & ---                           & ---                           & ---                                & ---       & ---     \\
    \bottomrule
  \end{tabular}
\end{table}

The deployment recipes in \Cref{tab:recipes} are short. For instruction-following and single-tool work (A0/A), a sub-5\,B model is sufficient at any target up to 90\%, at \$0.00007 to \$0.0010 per passed task on a Mac. For sequential 2-tool chains (B), a 2\,B model suffices at the 70\% reliability bar; no open-weight model in our corpus reaches the 80\% bar. C, D, and E are intervention territory (\Cref{sec:gap}); deployment there requires accepting reduced reliability, applying a model-specific intervention (\Cref{sec:interventions}), or routing to a frontier API.

\subsection{Cost and latency at matched accuracy}
\label{sec:cost}

\begin{figure}[tb]
  \centering
  \includegraphics[width=\linewidth]{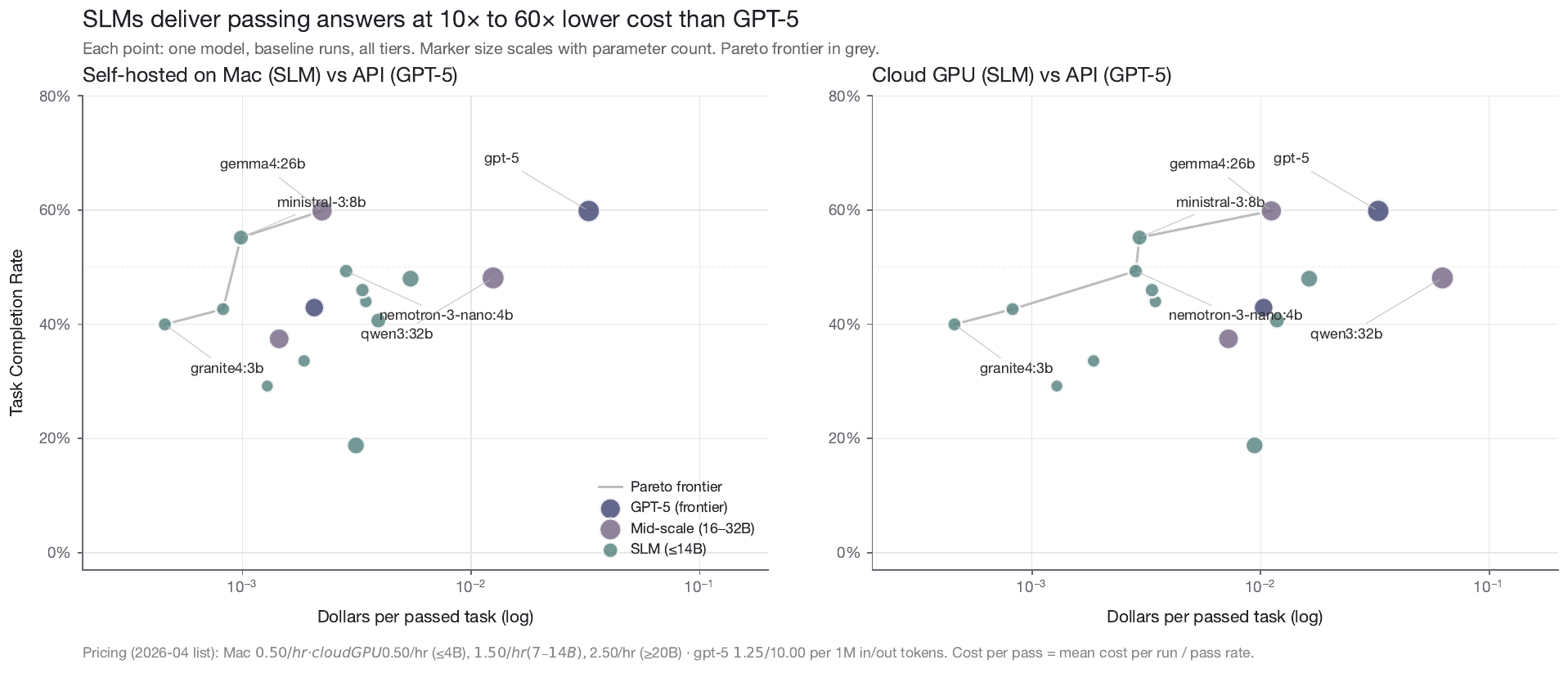}
  \caption{Cost-per-passed-task vs aggregate TCR. The Pareto frontier in this corpus is occupied by open-weight models at every reliability point; GPT-5 sits in the upper right. Pricing follows the locked 2026-04 schedule (\Cref{sec:methods}).}
  \label{fig:cost}
\end{figure}

\begin{figure}[tb]
  \centering
  \includegraphics[width=\linewidth]{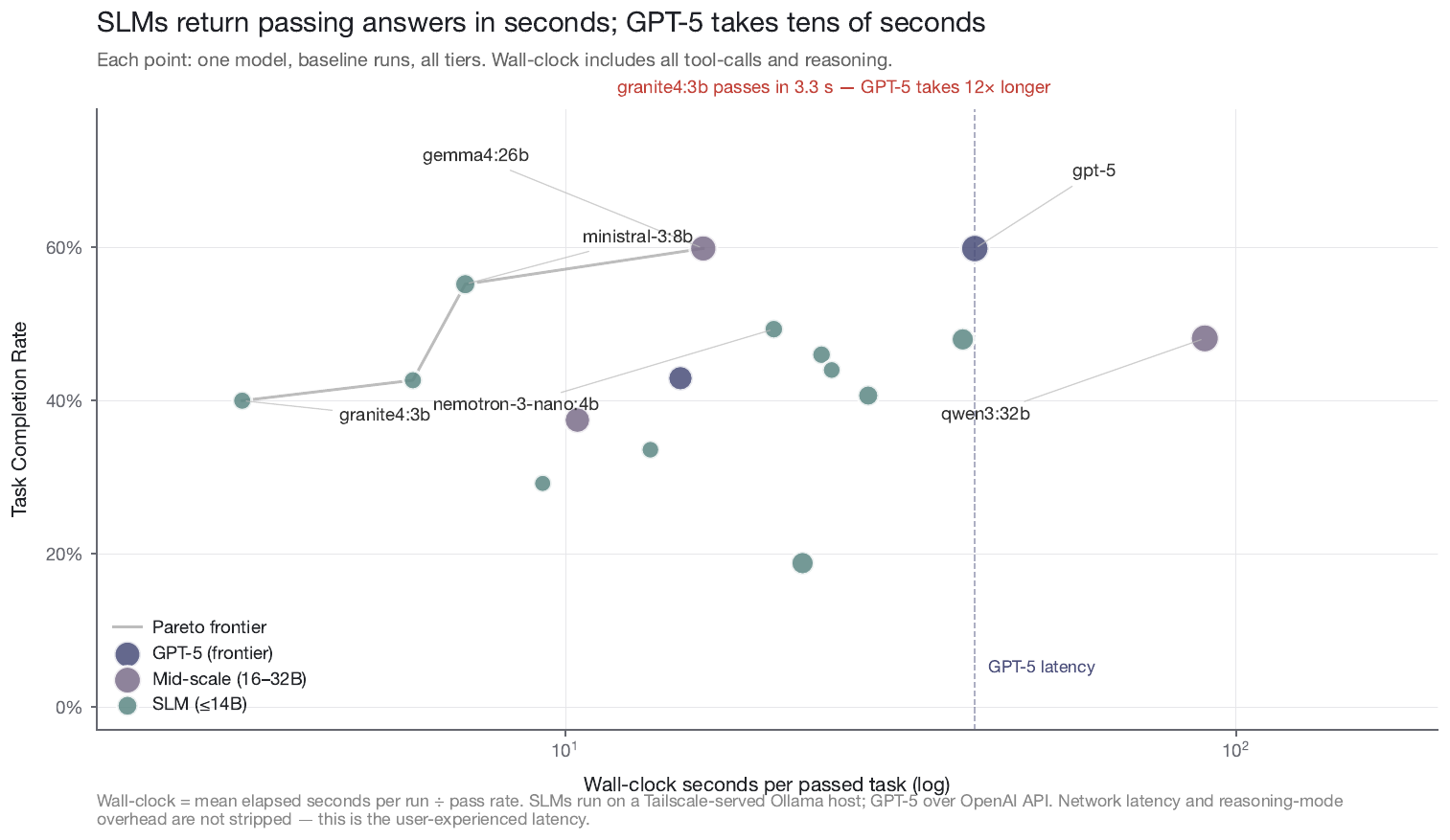}
  \caption{Wall-clock latency per passed task vs aggregate TCR. The \textit{qwen3:32b} cell with reasoning enabled (rightmost outlier) is slower per passed task than GPT-5 in this corpus.}
  \label{fig:latency}
\end{figure}

\Cref{fig:cost} (cost) and \Cref{fig:latency} (latency) put both axes on Pareto plots; \Cref{tab:cost} and \Cref{tab:latency} give the underlying numbers.

\begin{table}[ht]
  \centering
  \footnotesize
  \setlength{\tabcolsep}{3.5pt}
  \caption{Cost per passed task. Bold = cheapest reliable option (positive pass rate). All costs are shown in USD. Pricing assumptions are given in \Cref{sec:methods}.}
  \label{tab:cost}
  \begin{tabular}{@{}llrrrrrr@{}}
    \toprule
    Model & Params & TCR &
    \shortstack{\$/run\\(Mac)} &
    \shortstack{\$/run\\(GPU)} &
    \shortstack{\$/run\\(API)} &
    \shortstack{\$/pass\\(Mac/API)} &
    \shortstack{\$/pass\\(GPU/API)} \\
    \midrule
    \textit{functiongemma:270m}   & 0.27\,B & 0.0\%  & \$0.00006 & \$0.00006 & ---      & ---                  & ---                  \\
    \textit{qwen3:0.6b}           & 0.6\,B  & 29.2\% & \$0.00037 & \$0.00037 & ---      & \$0.0013             & \$0.0013             \\
    \textit{qwen3:1.7b}           & 1.7\,B  & 33.6\% & \$0.00062 & \$0.00062 & ---      & \$0.0019             & \$0.0019             \\
    \textit{qwen3.5:2b}           & 2\,B    & 44.0\% & \$0.0015  & \$0.0015  & ---      & \$0.0035             & \$0.0035             \\
    \textit{granite4:3b}          & 3\,B    & 40.0\% & \$0.00018 & \$0.00018 & ---      & \textbf{\$0.00046}   & \textbf{\$0.00046}   \\
    \textit{ministral-3:3b}       & 3\,B    & 42.7\% & \$0.00035 & \$0.00035 & ---      & \$0.00082            & \$0.00082            \\
    \textit{gemma4:e4b}           & 4\,B    & 46.0\% & \$0.0015  & \$0.0015  & ---      & \$0.0033             & \$0.0033             \\
    \textit{nemotron-3-nano:4b}   & 4\,B    & 49.3\% & \$0.0014  & \$0.0014  & ---      & \$0.0028             & \$0.0028             \\
    \textit{ministral-3:8b}       & 8\,B    & 55.2\% & \$0.00054 & \$0.0016  & ---      & \$0.00098            & \$0.0030             \\
    \textit{qwen3:8b}             & 8\,B    & 40.7\% & \$0.0016  & \$0.0048  & ---      & \$0.0039             & \$0.012              \\
    \textit{ministral-3:14b}      & 14\,B   & 18.8\% & \$0.00059 & \$0.0018  & ---      & \$0.0031             & \$0.0094             \\
    \textit{qwen3:14b}            & 14\,B   & 48.0\% & \$0.0026  & \$0.0078  & ---      & \$0.0054             & \$0.016              \\
    \textit{gpt-oss:20b}          & 20\,B   & 42.9\% & \$0.00088 & \$0.0044  & ---      & \$0.0021             & \$0.010              \\
    \textit{mistral-small3.2:24b} & 24\,B   & 37.5\% & \$0.00054 & \$0.0027  & ---      & \$0.0014             & \$0.0072             \\
    \textit{gemma4:26b}           & 26\,B   & 59.9\% & \$0.0013  & \$0.0067  & ---      & \$0.0022             & \$0.011              \\
    \textit{qwen3:32b}            & 32\,B   & 48.1\% & \$0.0060  & \$0.030   & ---      & \$0.012              & \$0.062              \\
    GPT-5                         & ---     & 59.9\% & ---       & ---       & \$0.020  & \$0.033              & \$0.033              \\
    \bottomrule
  \end{tabular}
\end{table}

GPT-5 sits at \$0.0327 per passed task at posted rates. Across the open-weight corpus, the cheapest reliable cell is \textit{granite4:3b} at \$0.00046 per pass on Mac --- 71$\times$ cheaper than GPT-5 --- at 40\% aggregate TCR. At matched 60\% aggregate TCR, \textit{gemma4:26b} is 15$\times$ cheaper on Mac and 3$\times$ cheaper on cloud GPU. \textit{Ministral-3:8b} sits between them at \$0.00098 per pass on Mac (33$\times$ cheaper) and 55\% aggregate TCR.

\begin{table}[tb]
  \centering
  \footnotesize
  \caption{Wall-clock latency per passed task. Bold = fastest reliable option. SLMs run on a Tailscale-served Ollama host (Mac); GPT-5 over the OpenAI API.}
  \label{tab:latency}
  \begin{tabular}{llrrr}
    \toprule
    Model & Params & TCR & Mean wall-clock per run & Wall-clock per passed task \\
    \midrule
    \textit{functiongemma:270m}   & 0.27\,B & 0.0\%  & 0.4\,s  & ---             \\
    \textit{qwen3:0.6b}           & 0.6\,B  & 29.2\% & 2.7\,s  & 9.2\,s          \\
    \textit{qwen3:1.7b}           & 1.7\,B  & 33.6\% & 4.5\,s  & 13.4\,s         \\
    \textit{qwen3.5:2b}           & 2\,B    & 44.0\% & 11.0\,s & 24.9\,s         \\
    \textit{granite4:3b}          & 3\,B    & 40.0\% & 1.3\,s  & \textbf{3.3\,s} \\
    \textit{ministral-3:3b}       & 3\,B    & 42.7\% & 2.5\,s  & 5.9\,s          \\
    \textit{gemma4:e4b}           & 4\,B    & 46.0\% & 11.1\,s & 24.1\,s         \\
    \textit{nemotron-3-nano:4b}   & 4\,B    & 49.3\% & 10.1\,s & 20.4\,s         \\
    \textit{ministral-3:8b}       & 8\,B    & 55.2\% & 3.9\,s  & 7.1\,s          \\
    \textit{qwen3:8b}             & 8\,B    & 40.7\% & 11.5\,s & 28.3\,s         \\
    \textit{ministral-3:14b}      & 14\,B   & 18.8\% & 4.2\,s  & 22.6\,s         \\
    \textit{qwen3:14b}            & 14\,B   & 48.0\% & 18.8\,s & 39.1\,s         \\
    \textit{gpt-oss:20b}          & 20\,B   & 42.9\% & 6.4\,s  & 14.8\,s         \\
    \textit{mistral-small3.2:24b} & 24\,B   & 37.5\% & 3.9\,s  & 10.4\,s         \\
    \textit{gemma4:26b}           & 26\,B   & 59.9\% & 9.6\,s  & 16.0\,s         \\
    \textit{qwen3:32b}            & 32\,B   & 48.1\% & 43.2\,s & 1.5\,min        \\
    GPT-5                         & ---     & 59.9\% & 24.4\,s & 40.8\,s         \\
    \bottomrule
  \end{tabular}
\end{table}

Wall-clock latency follows the cost direction. \textit{Granite4:3b} reaches 3.3\,s per passed task; \textit{ministral-3:8b} 7.1\,s; \textit{gemma4:26b} 16.1\,s; GPT-5 40.8\,s. At matched aggregate accuracy, \textit{gemma4:26b} is 2.5$\times$ faster than GPT-5. The deployment cautionary case is \textit{qwen3:32b} with reasoning enabled: it reaches 89.8\,s per passed task --- about 2.2$\times$ longer than GPT-5 --- at lower aggregate accuracy (48\% vs 61\%). On this configuration, enabling reasoning at the 32\,B scale leaves the model both slower and less accurate per passed task than GPT-5 in our corpus; \Cref{sec:interventions} reports a related finding on tier B.

On the cost-vs-TCR plane, \textit{granite4:3b} occupies the bottom-left, \textit{ministral-3:8b} and \textit{gemma4:26b} occupy the middle band, and GPT-5 sits alone in the upper right. The Pareto frontier of cost-per-passed-task is occupied by open-weight models at every reliability point we measure.

\section{Where the long-horizon gap moves}
\label{sec:gap}

\subsection{The gap is concentrated on E, and even there both sides struggle}
\label{sec:gap-on-E}

The \Cref{sec:results} results consolidate as follows. \textit{Gemma4:26b} is strictly higher than GPT-5 on A0, formally equivalent on A, descriptively close on B/C/D (point estimates within 9\,pp; per-tier paired sample size is insufficient for formal equivalence at the pre-registered $\pm 10$\,pp margin), and strictly inferior on E. Cost-per-passed-task is open-weight-dominant at every reliability point in our corpus. The one tier on which the frontier is strictly better is long-horizon planning under persistent constraints.

The E-tier tasks require the model to maintain and respect persistent constraints across 8--12 tool calls --- a step budget plus a constraint set that each tool call must check against. GPT-5's pass rate on E is 10\%; the highest open-weight cell on E is \textit{ministral-3:8b} at 16\% (CI overlapping GPT-5's, but failing Holm non-inferiority at any margin in our sweep). The frontier advantage on E is real and small in absolute terms; both sides are below practitioner-relevant reliability on this tier.

The next two subsections examine whether the E-tier gap responds to interventions that do not change parameter count, and why the interventions we tested do not generalize across models.

\subsection{Interventions are mechanism-specific}
\label{sec:interventions}

\begin{figure}[ht!]
  \centering
  \includegraphics[width=\linewidth]{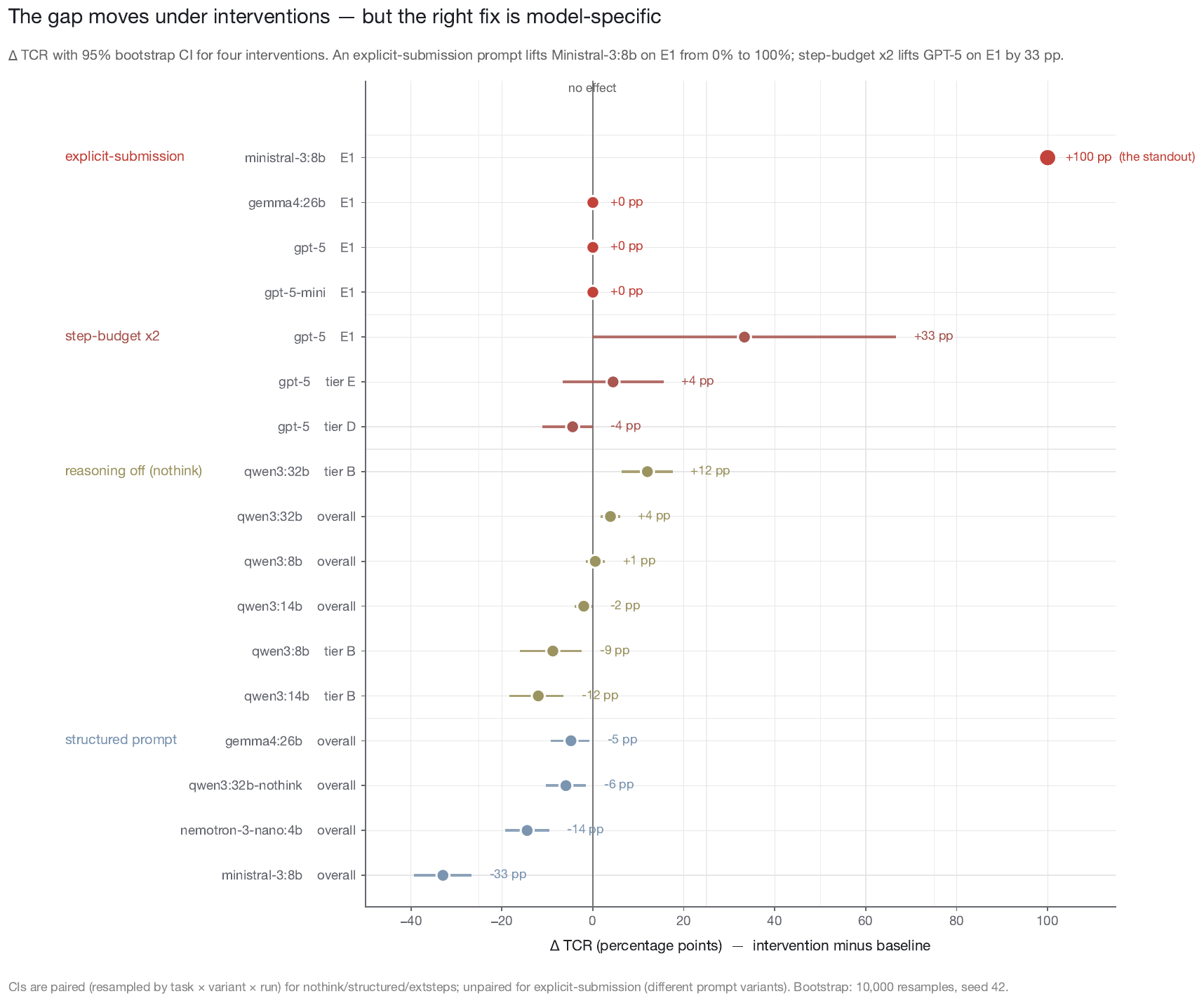}
  \caption{Targeted interventions on the long-horizon (E) gap and on Qwen3 reasoning mode (tier B). Bars show paired $\Delta$ TCR with 95\% bootstrap CIs over (task, variant, run\_idx) tuples. Interventions: explicit-submission system-prompt addition; doubled \textit{max\_steps} budget on D and E; reasoning-mode toggle on Qwen3; plan/execute/submit phase decomposition prompt.}
  \label{fig:interventions}
\end{figure}

\Cref{fig:interventions} plots the four largest effect sizes from our ablation sweep, with 95\% bootstrap CIs over paired (task, variant, run\_idx) tuples.

We ran four ablations on the residual long-horizon gap. The four divide cleanly into three model-specific lifts and one uniform regression.

\paragraph{Explicit-submission prompt: a $+100$\,pp lift on exactly one model.}
Adding ``you MUST call \textit{submit\_decision}'' to the system prompt moves \textit{ministral-3:8b} on E1 from 0/5 to 5/5 across two prompt variants --- a $+100$\,pp jump $[+100, +100]$. The same prompt is null on \textit{gemma4:26b}, \textit{qwen3:32b-nothink}, GPT-5, and GPT-5-mini, all of which remain at 0/5 on E1. Trace inspection clarifies the mechanism: \textit{ministral-3:8b} on E1 already executes a correct tool chain in canonical-prompt runs but emits a final-text answer without invoking \textit{submit\_decision}. The explicit prompt converts that text answer into a submission. The other four models fail before reaching the submission step, so the prompt has nothing to convert.

\paragraph{Step-budget $\times 2$: a $+33$\,pp effect localised to one task.}
Doubling GPT-5's \textit{max\_steps} budget on D and E moves only E1 (1/9 $\rightarrow$ 4/9, $+33$\,pp $[0, +67]$); the tier-E aggregate moves $+4$\,pp $[-7, +16]$. Termination-profile inspection confirms that GPT-5 never touches the extra step capacity on the other nine D/E tasks --- those failures are F5/F5b (resigns or plans without executing), not F4 (step exhaustion). E1 is the procurement-bundle task whose canonical solve path needs $\sim 8$ tool calls; the declared 10-step budget is genuinely tight there, and only there.

\paragraph{Reasoning mode is family-stratified.}
Disabling reasoning on Qwen3 helps the 32\,B model on tier B ($+12$\,pp $[+6, +18]$) but hurts the 8\,B ($-9$\,pp $[-16, -2]$) and the 14\,B ($-12$\,pp $[-18, -6]$) on the same tier. Tier B --- sequential 2-tool chaining --- is the uniquely reasoning-sensitive cell; A/C/D/E barely move. The reading we infer from traces is that at $\sim 14$\,B, reasoning tokens help compose a chain the model could not ``just know''; at $\sim 32$\,B the chain is already known and the reasoning tokens consume output budget that triggers additional F5 resignations. Reasoning mode is a deployment knob, not a capability axis.

\paragraph{Structured prompts hurt every model tested.}
A plan/execute/submit phase decomposition was the obvious candidate for a universal F5 reducer --- instruct the model to plan before executing and submit at the end. The data go the other way on every model in the sweep: \textit{gemma4:26b} $-5$\,pp $[-9, -1]$, \textit{qwen3:32b-nothink} $-6$\,pp, \textit{nemotron-3-nano:4b} $-14$\,pp $[-19, -10]$, \textit{ministral-3:8b} $-33$\,pp $[-39, -27]$ overall. Trace inspection of the largest regression (\textit{ministral-3:8b}) shows the model complying with the ``Do NOT call any tools yet'' instruction in the PLAN phase, then emitting a prose answer without ever entering EXECUTE. The structured prompt converts residual capability into a longer-form F5: a careful plan, no tool calls, a final-text answer. The intervention designed to reduce early resignation produces more of it.

The four ablations describe what moves and what does not. Explicit-submission addresses a latent-capability-without-required-action mode (\textit{ministral-3:8b} on E1); step-budget addresses a genuinely-tight-budget mode (GPT-5 on E1); reasoning-off addresses an over-reasoning mode (\textit{qwen3:32b} on B). Each effective intervention helped one model on one cell and was null on the others. The phase-decomposition prompt --- the obvious candidate for a universal lever --- regressed all four models we tried it on. We did not find a universal lever in this sweep.

\subsection{Failure-mode signatures explain why interventions don't generalize}
\label{sec:failures}

\begin{figure}[ht!]
  \centering
  \includegraphics[width=\linewidth]{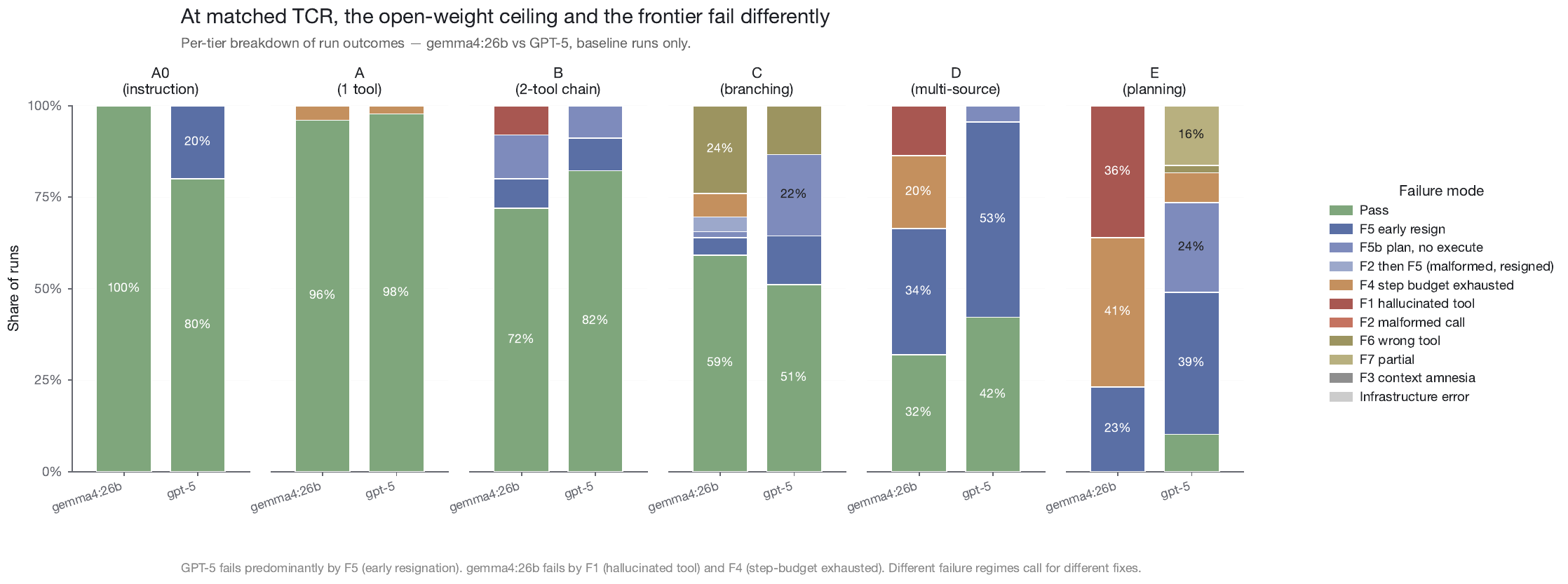}
  \caption{Tier-E failure-mode decomposition for GPT-5 ($n = 45$) and \textit{gemma4:26b} ($n = 125$). The two models reach aggregate TCR that is statistically indistinguishable at the pre-registered margin (\Cref{sec:parity}); their tier-E failure mass is concentrated in different categories. GPT-5: F5 (early resignation) 39\%, F5b (plan-without-execute) 24\%, F7 18\%, F4 8\%, F1 0\%. \textit{gemma4:26b}: F4 (step-budget exhausted) 41\%, F1 (hallucinated tool) 36\%, F5 23\%, F5b 0\%.}
  \label{fig:failures}
\end{figure}

\Cref{fig:failures} stacks the failure-mode decomposition for both models on tier E --- the cleanest contrast in the corpus.

GPT-5 and \textit{gemma4:26b} have aggregate TCR that is statistically indistinguishable at the pre-registered $\pm 10$\,pp margin (paired $\Delta = +0.4$\,pp, 90\% CI $[-4.0, +5.1]$; \Cref{sec:parity}). Their tier-E failure signatures are not.

\paragraph{GPT-5 on E ($n=45$).}
PASS 10\%, F5 39\%, F5b 24\%, F7 18\%, F4 8\%, F1 0\%. The frontier engages with E-tier tasks, plans, sometimes partially executes, then resigns (F5) or plans without executing (F5b). Step-budget exhaustion (F4) accounts for 4/45 runs and is concentrated entirely on E1 --- every other E task terminates well within budget. Hallucinated tool calls (F1) do not appear at all.

\paragraph{\textit{gemma4:26b} on E ($n=125$).}
PASS 0\%, F4 41\%, F1 36\%, F5 23\%. The mid-scale model exhausts its step budget on more than four runs in ten and invents tool names in another third --- the signature of a model running out of room and grasping for tools that don't exist. F5b is absent.

The two models reach indistinguishable overall TCR through categorically different mechanisms, and the asymmetry is what makes the \Cref{sec:interventions} interventions model-specific. GPT-5's E-tier failure mass is in F5/F5b, so a longer step budget can convert exhausted exploration into a successful submission (and does, on E1). \textit{Gemma4:26b}'s mass is in F4/F1, so a longer step budget would just give the model more room to invent more tools; the natural lever for an F1+F4 regime is a tool-template intervention or an explicit submission instruction earlier in the trajectory --- neither has been tested in this corpus. The same TCR cell hides different bottlenecks.

\section{Discussion}
\label{sec:discussion}

\paragraph{What the equivalence claim means.}
The aggregate-parity result (paired $\Delta = +0.4$\,pp, 90\% CI $[-4.0, +5.1]$ across 30 tasks) is a claim about \emph{this benchmark, this corpus, and this protocol}, with a margin chosen before analysis. Within those bounds it is a stronger claim than the literature standardly supports for agentic-tool-use benchmarks, where pass-rate aggregates usually go without confidence intervals at all. Outside those bounds --- different task distributions, different inference protocols, different frontier models --- the claim is not portable, and we do not portably make it. The findings we are willing to defend formally are: open-weight strict superiority on A0; formal equivalence on A; aggregate equivalence at the pre-registered $\pm 10$\,pp margin (the same data certify equivalence at any margin $\geq 5.1$\,pp); strict frontier superiority on E. The B/C/D verdicts are descriptive, not formal: per-tier sample size is insufficient to certify equivalence at the pre-registered margin even though point estimates are within 9\,pp of GPT-5 on all three tiers.

\paragraph{What this implies for routing.}
The Pareto frontier of cost-per-passed-task is occupied by open-weight models at every reliability point in our corpus. For tiers A0, A, and B at the 80\% reliability bar (with B requiring the relaxed 70\% bar), the deployment recipe is sub-5\,B and Mac-hostable, at \$0.00007 to \$0.0010 per passed task (\Cref{tab:recipes}). For C, D, and E, no zero-shot configuration in our corpus clears even the 60\% bar; deployment there requires accepting reduced reliability, applying a model-specific intervention (\Cref{sec:interventions}), or routing to a frontier API. The asymmetry between the dense open-weight Pareto frontier on cost and latency and the empty cells on C/D/E marks the practitioner choice: route the lower tiers to small open-weight models, and reserve the frontier for the tier where it is needed --- long-horizon planning under persistent constraints.

\paragraph{Three observations we do not fully understand.}
Three cells in the corpus are descriptively striking and resist clean explanation. The highest tier-D cell is a 4\,B model (\textit{nemotron-3-nano:4b}, 36\% [28, 44]), not the 26\,B mid-scale (\textit{gemma4:26b}, 32\%): multi-source synthesis with recovery does not scale monotonically with parameter count in this corpus. \textit{Ministral-3:14b} is dominated cell by cell by its 8\,B sibling (A0 49 vs 75; A 24 vs 84; B 28 vs 84) --- a family-internal regression of large magnitude. \textit{Mistral-small3.2:24b} reaches 96\% on A0 and 93\% on A but collapses to 16\% on B; an SDR of 28.6\% on B is consistent with a tool-template incompatibility on the native tool-calling path rather than a capability ceiling. These suggest that parameter count alone is a poor predictor of agentic capability across model families.

\paragraph{What we would want to see next.}
A second flagship anchor (Anthropic or Google) to cross-validate the per-tier verdicts; more E-tier task designs to broaden the long-horizon picture beyond five tasks; an informal human-pass baseline on a sampled subset to put the C/D/E cells in context; and targeted F5-mass decomposition prompts on a wider model set to test whether the explicit-submission lift on \textit{ministral-3:8b} is one of many latent-capability cells or a singular instance.

\section{Limitations}
\label{sec:limitations}

Our conclusions are specific to this benchmark, protocol, and model set. GPT-5 is the only frontier model evaluated at paper-grade paired sample size, so the aggregate comparison is well powered, but the per-tier paired analyses on B, C, and D remain too underpowered for formal equivalence at the pre-registered margin. Each tier also contains only five tasks, which is sufficient for a controlled capability study but not for exhaustive coverage of agentic workloads. In addition, AgentFloor measures native tool-calling control in a deterministic abstract-tool environment rather than end-to-end performance in live settings with real APIs, web interaction, GUI grounding, or prompt-based tool emulation. Finally, we do not include a human baseline, and some failure mass, especially early resignation, may reflect compliance or prompting effects rather than hard capability limits. The higher-tier ceiling should therefore be interpreted as a controlled estimate under this evaluation setup, not as a universal limit on open-weight or frontier models.

\section{Conclusion}
\label{sec:conclusion}

We have reported a controlled 30-task tool use ladder run against 16 open-weight models and GPT-5. At the workload level (30 tasks paired, $n = 270$), \textit{gemma4:26b} is equivalent to GPT-5 at the pre-registered $\pm 10$\,pp margin (paired $\Delta = +0.4$\,pp, 90\% CI $[-4.0, +5.1]$) while running roughly 15$\times$ cheaper on Mac self-hosting (3$\times$ on cloud GPU) and 2.5$\times$ faster per passed task. Open-weight is strictly higher on the no-tool tier and formally equivalent on single-tool use; on B, C, and D the per-tier paired sample size does not certify equivalence at the pre-registered margin even though point estimates are within 9\,pp; on long-horizon planning under persistent constraints, GPT-5 is strictly better, and neither side reaches practitioner-relevant reliability. Some of the residual long-horizon gap moves under targeted interventions, but each effective intervention we tested helped one model and was null on others; the obvious universal candidate --- a plan/execute/submit phase decomposition prompt --- regressed every model we tried it on. We release the benchmark, the harness, the sweep configurations, and the full 16{,}542-run corpus.

\bibliographystyle{plainnat}
\bibliography{refs}

@article{zhou2023webarena,
  title={Webarena: A realistic web environment for building autonomous agents},
  author={Zhou, Shuyan and Xu, Frank F and Zhu, Hao and Zhou, Xuhui and Lo, Robert and Sridhar, Abishek and Cheng, Xianyi and Ou, Tianyue and Bisk, Yonatan and Fried, Daniel and others},
  journal={arXiv preprint arXiv:2307.13854},
  year={2023}
}

@article{zhong2025complexfuncbench,
  title={ComplexFuncBench: Exploring multi-step and constrained function calling under long-context scenario},
  author={Zhong, Lucen and Du, Zhengxiao and Zhang, Xiaohan and Hu, Haiyi and Tang, Jie},
  journal={arXiv preprint arXiv:2501.10132},
  year={2025}
}

@article{yang2025qwen3,
  title={Qwen3 technical report},
  author={Yang, An and Li, Anfeng and Yang, Baosong and Zhang, Beichen and Hui, Binyuan and Zheng, Bo and Yu, Bowen and Gao, Chang and Huang, Chengen and Lv, Chenxu and others},
  journal={arXiv preprint arXiv:2505.09388},
  year={2025}
}

@article{wang2023mint,
  title={Mint: Evaluating llms in ƒmulti-turn interaction with tools and language feedback},
  author={Wang, Xingyao and Wang, Zihan and Liu, Jiateng and Chen, Yangyi and Yuan, Lifan and Peng, Hao and Ji, Heng},
  journal={arXiv preprint arXiv:2309.10691},
  year={2023}
}

@article{ong2024routellm,
  title={Routellm: Learning to route llms with preference data},
  author={Ong, Isaac and Almahairi, Amjad and Wu, Vincent and Chiang, Wei-Lin and Wu, Tianhao and Gonzalez, Joseph E and Kadous, M Waleed and Stoica, Ion},
  journal={arXiv preprint arXiv:2406.18665},
  year={2024}
}

@article{aggarwal2023automix,
  title={Automix: Automatically mixing language models},
  author={Aggarwal, Pranjal and Madaan, Aman and Anand, Ankit and Potharaju, Srividya Pranavi and Mishra, Swaroop and Zhou, Pei and Gupta, Aditya and Rajagopal, Dheeraj and Kappaganthu, Karthik and Yang, Yiming and others},
  journal={arXiv preprint arXiv:2310.12963},
  year={2023}
}

@inproceedings{abdelaziz2024granite,
  title={Granite-function calling model: Introducing function calling abilities via multi-task learning of granular tasks},
  author={Abdelaziz, Ibrahim and Basu, Kinjal and Agarwal, Mayank and Kumaravel, Sadhana and Stallone, Matthew and Panda, Rameswar and Rizk, Yara and Bhargav, GP Shrivatsa and Crouse, Maxwell and Gunasekara, Chulaka and others},
  booktitle={Proceedings of the 2024 Conference on Empirical Methods in Natural Language Processing: Industry Track},
  pages={1131--1139},
  year={2024}
}

@article{garg2025real,
  title={Real: Benchmarking autonomous agents on deterministic simulations of real websites},
  author={Garg, Divyansh and VanWeelden, Shaun and Caples, Diego and Draguns, Andis and Ravi, Nikil and Putta, Pranav and Garg, Naman and Abraham, Tomas and Lara, Michael and Lopez, Federico and others},
  journal={arXiv preprint arXiv:2504.11543},
  year={2025}
}

@article{grattafiori2024llama,
  title={The llama 3 herd of models},
  author={Grattafiori, Aaron and Dubey, Abhimanyu and Jauhri, Abhinav and Pandey, Abhinav and Kadian, Abhishek and Al-Dahle, Ahmad and Letman, Aiesha and Mathur, Akhil and Schelten, Alan and Vaughan, Alex and others},
  journal={arXiv preprint arXiv:2407.21783},
  year={2024}
}

@inproceedings{guo2024stabletoolbench,
  title={Stabletoolbench: Towards stable large-scale benchmarking on tool learning of large language models},
  author={Guo, Zhicheng and Cheng, Sijie and Wang, Hao and Liang, Shihao and Qin, Yujia and Li, Peng and Liu, Zhiyuan and Sun, Maosong and Liu, Yang},
  booktitle={Findings of the Association for Computational Linguistics: ACL 2024},
  pages={11143--11156},
  year={2024}
}

@article{jimenez2023swe,
  title={Swe-bench: Can language models resolve real-world github issues?},
  author={Jimenez, Carlos E and Yang, John and Wettig, Alexander and Yao, Shunyu and Pei, Kexin and Press, Ofir and Narasimhan, Karthik},
  journal={arXiv preprint arXiv:2310.06770},
  year={2023}
}

@article{kokane2024toolscan,
  title={Toolscan: A benchmark for characterizing errors in tool-use llms},
  author={Kokane, Shirley and Zhu, Ming and Awalgaonkar, Tulika and Zhang, Jianguo and Hoang, Thai and Prabhakar, Akshara and Liu, Zuxin and Lan, Tian and Yang, Liangwei and Tan, Juntao and others},
  journal={arXiv preprint arXiv:2411.13547},
  year={2024}
}

@article{liu2023agentbench,
  title={Agentbench: Evaluating llms as agents},
  author={Liu, Xiao and Yu, Hao and Zhang, Hanchen and Xu, Yifan and Lei, Xuanyu and Lai, Hanyu and Gu, Yu and Ding, Hangliang and Men, Kaiwen and Yang, Kejuan and others},
  journal={arXiv preprint arXiv:2308.03688},
  year={2023}
}

@misc{patil2023gorilla,
  title         = {Gorilla: Large Language Model Connected with Massive APIs},
  author        = {Patil, Shishir G. and Zhang, Tianjun and Wang, Xin and Gonzalez, Joseph E.},
  year          = {2023},
  eprint        = {2305.15334},
  archivePrefix = {arXiv},
  primaryClass  = {cs.CL},
  url           = {https://arxiv.org/abs/2305.15334}
}

@misc{li2023apibank,
  title         = {API-Bank: A Comprehensive Benchmark for Tool-Augmented LLMs},
  author        = {Li, Minghao and Zhao, Yingxiu and Yu, Bowen and Song, Feifan and Li, Hangyu and Yu, Haiyang and Li, Zhoujun and Huang, Fei and Li, Yongbin},
  year          = {2023},
  eprint        = {2304.08244},
  archivePrefix = {arXiv},
  primaryClass  = {cs.CL},
  url           = {https://arxiv.org/abs/2304.08244}
}

@misc{patil2024bfcl,
  title        = {Berkeley Function Calling Leaderboard ({BFCL})},
  author       = {{Berkeley Gorilla Team}},
  year         = {2024},
  howpublished = {\url{https://gorilla.cs.berkeley.edu/leaderboard.html}},
  note         = {Accessed 2026-04-27}
}

@misc{yao2024taubench,
  title         = {{$\tau$-bench: A Benchmark for Tool-Agent-User Interaction in Real-World Domains}},
  author        = {Yao, Shunyu and Shinn, Noah and Razavi, Pedram and Narasimhan, Karthik},
  year          = {2024},
  eprint        = {2406.12045},
  archivePrefix = {arXiv},
  primaryClass  = {cs.AI},
  url           = {https://arxiv.org/abs/2406.12045}
}

@misc{yuan2025critictool,
  title         = {CRITICTOOL: Evaluating Self-Critique Capabilities of Large Language Models in Tool-Calling Error Scenarios},
  author        = {Huang, Shiting and Fang, Zhen and Chen, Zehui and Yuan, Siyu and Ye, Junjie and Zeng, Yu and Chen, Lin and Mao, Qi and Zhao, Feng},
  year          = {2025},
  eprint        = {2506.13977},
  archivePrefix = {arXiv},
  primaryClass  = {cs.CL},
  url           = {https://arxiv.org/abs/2506.13977}
}

@misc{ruan2023toolemu,
  title         = {ToolEmu: Emulating Large Language Model Agents for Safe and Effective Tool Learning},
  author        = {Ruan, Yangjun and others},
  year          = {2023},
  eprint        = {2309.15817},
  archivePrefix = {arXiv},
  primaryClass  = {cs.CL},
  url           = {https://arxiv.org/abs/2309.15817}
}

@misc{ma2024agentboard,
  title         = {AgentBoard: An Analytical Evaluation Board of Multi-turn LLM Agents},
  author        = {Ma, Chang and Zhang, Junlei and Zhu, Zhihao and Yang, Cheng and Yang, Yujiu and Jin, Yaohui and Lan, Zhenzhong and Kong, Lingpeng and He, Junxian},
  year          = {2024},
  eprint        = {2401.13178},
  archivePrefix = {arXiv},
  primaryClass  = {cs.AI},
  url           = {https://arxiv.org/abs/2401.13178}
}

@misc{xie2024osworld,
  title         = {OSWorld: Benchmarking Multimodal Agents for Open-Ended Tasks in Real Computer Environments},
  author        = {Xie, Tianbao and Zhang, Danyang and Chen, Jixuan and Li, Xiaochuan and Zhao, Siheng and Cao, Ruisheng and Hua, Toh Jing and Cheng, Zhoujun and Shin, Dongchan and Lei, Fangyu and others},
  year          = {2024},
  eprint        = {2404.07972},
  archivePrefix = {arXiv},
  primaryClass  = {cs.AI},
  url           = {https://arxiv.org/abs/2404.07972}
}

@misc{mialon2023gaia,
  title         = {GAIA: a Benchmark for General AI Assistants},
  author        = {Mialon, Gr{\'e}goire and Fourrier, Cl{\'e}mentine and Swift, Craig and Wolf, Thomas and LeCun, Yann and Scialom, Thomas},
  year          = {2023},
  eprint        = {2311.12983},
  archivePrefix = {arXiv},
  primaryClass  = {cs.AI},
  url           = {https://arxiv.org/abs/2311.12983}
}

@misc{lu2024slmsurvey,
  title         = {Small Language Models: Survey, Measurements, and Insights},
  author        = {Lu, Zhenyan and Li, Xiang and Cai, Dongqi and Yi, Rongjie and Liu, Fangming and Zhang, Xiwen and Lane, Nicholas D. and Xu, Mengwei},
  year          = {2024},
  eprint        = {2409.15790},
  archivePrefix = {arXiv},
  primaryClass  = {cs.CL},
  url           = {https://arxiv.org/abs/2409.15790}
}

@misc{abdin2024phi3,
  title         = {Phi-3 Technical Report: A Highly Capable Language Model Locally on Your Phone},
  author        = {Abdin, Marah and Jacobs, Sam and Awan, Ammar and Aneja, Jyoti and Awadalla, Hany and others},
  year          = {2024},
  eprint        = {2404.14219},
  archivePrefix = {arXiv},
  primaryClass  = {cs.CL},
  url           = {https://arxiv.org/abs/2404.14219}
}

@misc{gemma2024report,
  title         = {Gemma: Open Models Based on Gemini Research and Technology},
  author        = {{Gemma Team} and others},
  year          = {2024},
  eprint        = {2403.08295},
  archivePrefix = {arXiv},
  primaryClass  = {cs.CL},
  url           = {https://arxiv.org/abs/2403.08295}
}

@misc{mahmood2025smallmodelsbigtasks,
  title         = {Small Models, Big Tasks: An Exploratory Empirical Study on Small Language Models for Function Calling},
  author        = {Kavathekar, Ishan and Donakanti, Raghav and Kumaraguru, Ponnurangam and Vaidhyanathan, Karthik},
  year          = {2025},
  eprint        = {2504.19277},
  archivePrefix = {arXiv},
  primaryClass  = {cs.CL},
  url           = {https://arxiv.org/abs/2504.19277}
}

@misc{belcak2025slmsfutureagentic,
  title         = {Small Language Models are the Future of Agentic AI},
  author        = {Belcak, Peter and Heinrich, Greg and Diao, Shizhe and Fu, Yonggan and Dong, Xin and Muralidharan, Saurav and Lin, Yingyan Celine and Molchanov, Pavlo},
  year          = {2025},
  eprint        = {2506.02153},
  archivePrefix = {arXiv},
  primaryClass  = {cs.AI},
  url           = {https://arxiv.org/abs/2506.02153}
}

@misc{cemri2025mast,
  title         = {Why Do Multi-Agent LLM Systems Fail?},
  author        = {Cemri, Mert and Pan, Melissa Z. and Yang, Shuyi and Agrawal, Lakshya A. and Chopra, Bhavya and Tiwari, Rishabh and Keutzer, Kurt and Parameswaran, Aditya and Klein, Dan and Ramchandran, Kannan and others},
  year          = {2025},
  eprint        = {2503.13657},
  archivePrefix = {arXiv},
  primaryClass  = {cs.AI},
  url           = {https://arxiv.org/abs/2503.13657}
}

@misc{chen2023frugalgpt,
  title         = {FrugalGPT: How to Use Large Language Models While Reducing Cost and Improving Performance},
  author        = {Chen, Lingjiao and Zaharia, Matei and Zou, James},
  year          = {2023},
  eprint        = {2305.05176},
  archivePrefix = {arXiv},
  primaryClass  = {cs.LG},
  url           = {https://arxiv.org/abs/2305.05176}
}

@misc{ding2024hybridllm,
  title         = {Hybrid LLM: Cost-Efficient and Quality-Aware Query Routing},
  author        = {Ding, Dujian and Mallick, Ankur and Wang, Chi and Sim, Robert and Mukherjee, Subhabrata and Ruhle, Victor and Lakshmanan, Laks V. S. and Awadallah, Ahmed Hassan},
  year          = {2024},
  eprint        = {2404.14618},
  archivePrefix = {arXiv},
  primaryClass  = {cs.LG},
  url           = {https://arxiv.org/abs/2404.14618}
}

\end{document}